\documentclass[letterpaper]{article} 
\usepackage{aaai24}  
\usepackage{times}  
\usepackage{helvet}  
\usepackage{courier}  
\usepackage[hyphens]{url}  
\usepackage{graphicx} 
\usepackage{natbib}  
\usepackage{caption} 
\usepackage{algorithm}
\usepackage{algorithmic}

\newtheorem{definition}{Definition}

\usepackage{enumerate}
\usepackage{multirow}
\usepackage{bbding}
\usepackage[table,xcdraw]{xcolor}

\usepackage{subfig}

\usepackage{graphicx}
\usepackage{amssymb}
\usepackage{bm}
\usepackage{booktabs}

\usepackage{amsmath}

\usepackage{newfloat}
\usepackage{listings}
\DeclareCaptionStyle{ruled}{labelfont=normalfont,labelsep=colon,strut=off} 
\floatstyle{ruled}
\newfloat{listing}{tb}{lst}{}
\floatname{listing}{Listing}
\title{ADA-GAD: Anomaly-Denoised Autoencoders for Graph Anomaly Detection}

\author {
Junwei He\textsuperscript{\rm 1,2}
Qianqian Xu\textsuperscript{\rm 1}\thanks{Corresponding Authors}, 
Yangbangyan Jiang\textsuperscript{\rm 2}, 
Zitai Wang\textsuperscript{\rm 3,4}, 
Qingming Huang\textsuperscript{\rm 1,2,5$*$}\\
}
\affiliations {
\textsuperscript{\rm 1}Key Laboratory of Intelligent Information Processing, Institute of Computing Technology, CAS, Beijing, China \\
\textsuperscript{\rm 2}School of Computer Science and Technology, University of Chinese Academy of Sciences, Beijing, China \\
\textsuperscript{\rm 3}Institute of Information Engineering, CAS, Beijing, China \\
\textsuperscript{\rm 4}School of Cyber Security, University of Chinese Academy of Sciences, Beijing, China \\
\textsuperscript{\rm 5}Key Laboratory of Big Data Mining and Knowledge Management, Chinese Academy of Sciences, Beijing, China \\
\{hejunwei22s, xuqianqian\}@ict.ac.cn, jiangyangbangyan@ucas.ac.cn, wangzitai@iie.ac.cn, qmhuang@ucas.ac.cn \\
}

\usepackage{bibentry}

\begin{document}
\maketitle
\begin{abstract}
Graph anomaly detection is crucial for identifying nodes that deviate from regular behavior within graphs, benefiting various domains such as fraud detection and social network. Although existing reconstruction-based methods have achieved considerable success, they may face the \textit{Anomaly Overfitting} and \textit{Homophily Trap} problems caused by the abnormal patterns in the graph, breaking the assumption that normal nodes are often better reconstructed than abnormal ones. Our observations indicate that models trained on graphs with fewer anomalies exhibit higher detection performance. Based on this insight, we introduce a novel two-stage framework called Anomaly-Denoised Autoencoders for Graph Anomaly Detection (ADA-GAD). In the first stage, we design a learning-free anomaly-denoised augmentation method to generate graphs with reduced anomaly levels. We pretrain graph autoencoders on these augmented graphs at multiple levels, which enables the graph autoencoders to capture normal patterns. In the next stage, the decoders are retrained for detection on the original graph, benefiting from the multi-level representations learned in the previous stage. Meanwhile, we propose the node anomaly distribution regularization to further alleviate \textit{Anomaly Overfitting}. We validate the effectiveness of our approach through extensive experiments on both synthetic and real-world datasets.

\end{abstract}

\section{Introduction}

\begin{figure}[t]
    \centering
    \subfloat[Previous Reconstruction-based GAD Framework]{\includegraphics[width=0.42\textwidth]{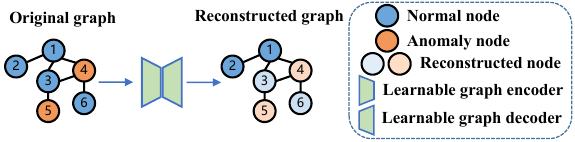}\label{fig:intro_a}}
    \\
    \subfloat[The Proposed Two-stage GAD Framework]{\includegraphics[width=0.43\textwidth]{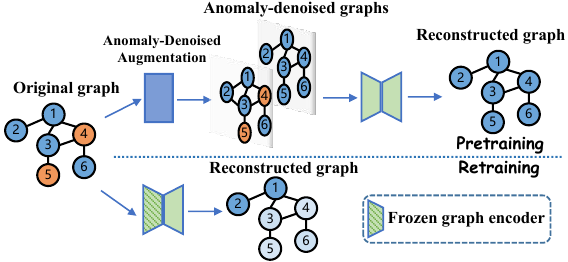}\label{fig:intro_b}}
    \caption{Workflow comparison. Previous reconstruction-based methods are trained on the contaminated graph. In contrast, our framework involves pretraining on anomaly-denoised graphs to reduce the impact of anomalous nodes.}\label{fig:intro_ab}
\end{figure}

The goal of unsupervised graph anomaly detection (GAD) is to identify rare patterns that deviate from the majority patterns in a graph, which has been extensively applied in diverse domains, such as fraud detection \cite{abdallah2016fraud,cheng2020graph,dou2020enhancing} and social network \cite{Fan_Zhang_Li_2020,duan2023graph}. Recently, reconstruction-based Graph Neural Networks (GNNs) methods have achieved great success and have become the mainstream approach. The common assumption is that normal nodes are easier to be reconstructed than abnormal nodes. On this basis, such methods usually train a graph autoencoder and determine anomalies according to the magnitude of the reconstruction errors.

However, the anomalous patterns in the graph might hinder the performance of reconstruction-based methods in two ways. (1) \textbf{Anomaly Overfitting}: graphs in the real world are highly sparse, and powerful GNNs tend to overfit to anomalous features, leading to small reconstruction errors even for anomalies. This, in turn, can cause the model to fail. (2) \textbf{Homophily Trap}: Most GNNs operate under the homophily assumption \cite{kipf2016semi}, which suggests that connected nodes share similar features. Therefore, the presence of anomalous nodes may hinder the reconstruction of nearby normal nodes, such that the corresponding magnified reconstruction errors bias the detection results. These phenomena are illustrated in Figure \ref{fig:intro_a}. The reconstructed features of normal nodes 3 and 6 are influenced by their anomalous neighbors 4 and 5 due to \textit{Homophily Trap}. Meanwhile, owing to \textit{Anomaly Overfitting}, nodes 4 and 5 are well-reconstructed, far from what we expected.

\begin{figure}[t]
    \centering
    \subfloat {\includegraphics[width=0.24\textwidth,height=0.17\textwidth]{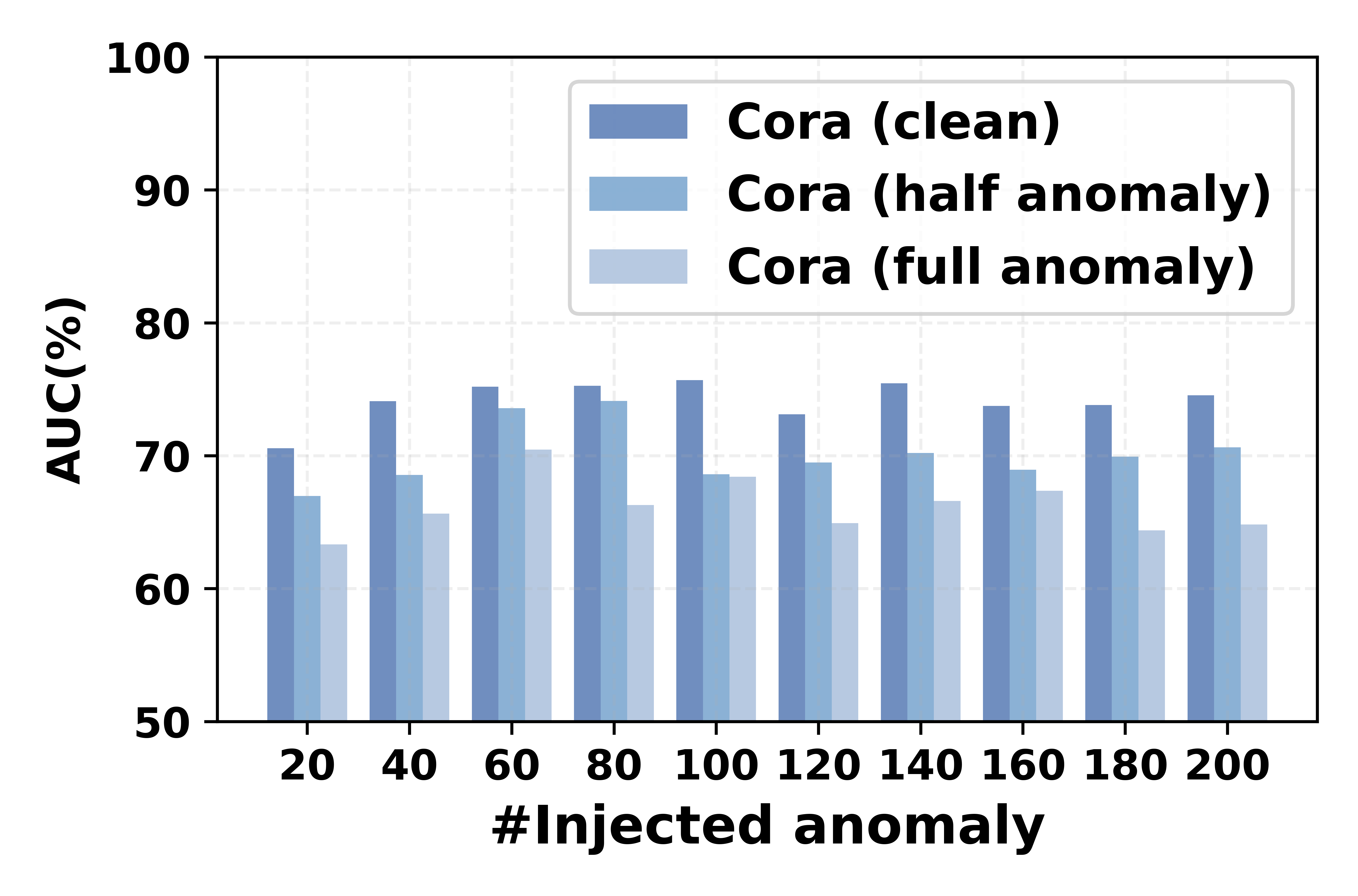}\label{fig:intro_1}}
    \subfloat {\includegraphics[width=0.24\textwidth,height=0.17\textwidth]{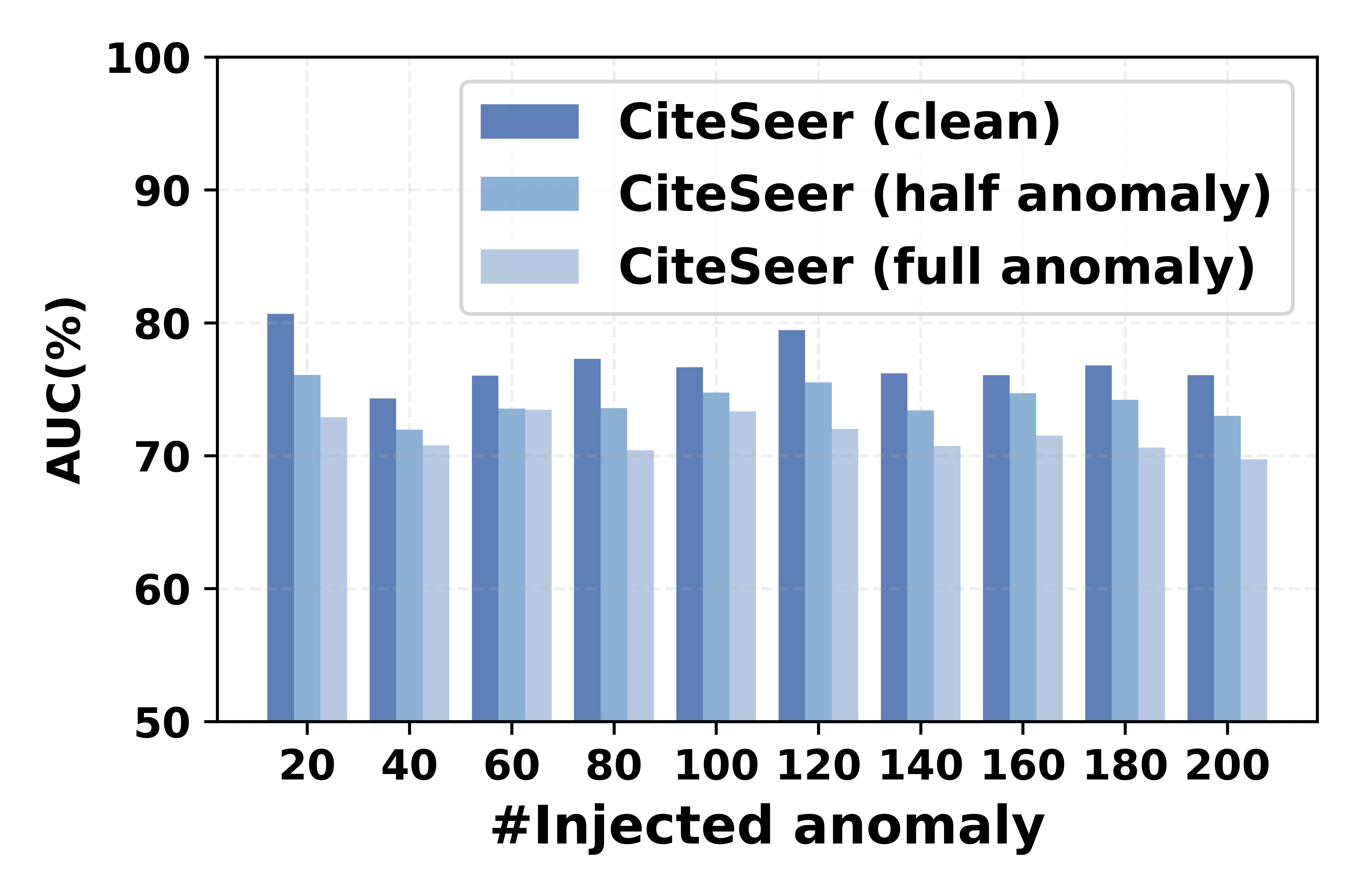}\label{fig:intro_2}}
    \caption{Detection performance of the DOMINANT model on Cora and CiteSeer datasets. The x-axis denotes the number of injected anomalies ($n$), while the y-axis shows the test results for models trained on graphs with clean (no anomalies), half-injected ($n/2$), or fully-injected ($n$) anomalies, but all evaluated on graphs containing $n$ injected anomalies.  We see that the less the training data is contaminated, the better the performance is.}\label{fig:intro_exp}

\end{figure}

We conduct a simple experiment to verify the negative effects of the anomalous patterns. Specifically, the popular DOMINANT baseline \cite{Ding_Li_Bhanushali_Liu_2019} are trained on the Cora and CiteSeer datasets \cite{sen2008collective} under three settings: on the original graph containing no abnormal nodes, on the partially-contaminated graph with $n/2$ injected anomalies, and on the graph with fully-injected $n$ anomalies. During the testing phase, all the models are tested on graphs with $n$ anomalies. As shown in Figure \ref{fig:intro_exp}, the model trained on the clean graph consistently outperforms the others under different numbers of injected anomalies. Moreover, even converting only half of the anomalies into clean data for training can improve performance. Namely, the less the training data is contaminated, the better the detection performance is.

\begin{figure}[!t]
  \centering
  \includegraphics[width=0.43\textwidth]{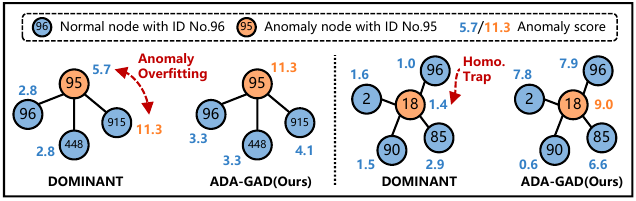}
\caption{Real cases of \textit{Anomaly Overfitting} and \textit{Homophily Trap} on Disney and Books datasets. Compared with DOMINANT, our ADA-GAD can effectively mitigate these issues.}

\label{fig:intro-c}
\end{figure}

Motivated by this, we hope to train the model on a graph as clean as possible. Since the ground-truth clean graph is not available, we need to find a way to reduce the anomaly level of the graph for training and effectively leverage the graph for detection.  To this aim, we present a two-stage framework called Anomaly-Denoised Autoencoders for Graph Anomaly Detection (ADA-GAD), as illustrated in Figure \ref{fig:intro_ab}(b). \textbf{(1) Stage 1:} We develop a learning-free augmentation method to obtain cleaner graphs, whose anomaly degree is quantified by a spectral property-based metric. Such anomaly-denoised augmentation technique generates three levels of augmented graphs by masking: node-level, edge-level, and subgraph-level. Corresponding anomaly-denoised autoencoders are pretrained on these augmented graphs using masking pretraining strategies, forcing the model to discover normal patterns. \textbf{(2) Stage 2:} We freeze the pretrained encoders and retrain the decoders from scratch to reconstruct the original graph for detection. We utilize an attention mechanism to aggregate the frozen multi-level representations, and introduce node anomaly distribution regularization, which sharpens the anomaly score distribution of nodes to prevent \textit{Anomaly Overfitting}. Subsequently, we identify anomalous nodes based on the magnitude of the reconstruction error. The efficacy of our ADA-GAD framework is visualized in Figure \ref{fig:intro-c}. In comparison with the previous methods, our ADA-GAD exhibits a significant reduction in the issues of \textit{Anomaly Overfitting} and \textit{Homophily Trap}.

The contributions of this paper are three-fold:
\begin{itemize}
    \item To alleviate the phenomena of \textit{Anomaly Overfitting} and \textit{Homophily Trap}, we propose a two-stage graph anomaly detection framework ADA-GAD that firstly reduces the anomaly level of the graph for pretraining and then retrains the decoder for detection.  
    \item In the pretraining stage, we present a metric to quantify the degree of anomaly in a graph. Then an anomaly-denoised augmentation strategy is introduced to generate augmented graphs with lower anomaly degrees for multi-level masking pretraining.
    \item In the retraining stage, we design a regularization term to make the distribution of each node's anomaly score sharper, especially to overcome the \textit{Anomaly Overfitting} issue.
\end{itemize}
Extensive experiments on two synthetic and five real-world anomaly datasets demonstrate the superiority of the proposed method.

\section{Related Work}

\paragraph{\textbf{Graph Neural Networks}} Graph neural networks (GNNs) are widely used in various deep learning tasks, as they can process graph-structured data and learn both the structural and attributive information of graphs \cite{kipf2016semi,velivckovic2017graph,gupta2021graph}, which have achieved remarkable results in tasks such as social networks, recommendation systems and bioinformatics \cite{zhou2020graph,waikhom2021graph}. GNNs can  be divided into two types: spectral-based and spatial-based \cite{zhu2021interpreting}. Spectral-based GNNs use spectral graph theory and rely on the Laplacian matrix of the graph, while spatial-based GNNs use the spatial information of the nodes and rely on message passing mechanisms \cite{kipf2016semi,xu2018powerful}. Typical spectral-based models include ChebNet \cite{defferrard2016convolutional} and GCN \cite{kipf2016semi}, while classical spatial-based GNNs are GAT \cite{velickovic2017graph}, GraphSAGE \cite{hamilton2017inductive}, GIN \cite{xu2018powerful}, and GraphSNN \cite{wijesinghe2022new}.

\paragraph{\textbf{Anomaly Detection on Static Attributed Graphs}}  Graph anomaly detection \cite{duan2023graph} aims to identify nodes that are different from most nodes. Some progress has been made in anomaly detection on static attributed graphs.
non-deep learning methods \cite{Li_Dani_Hu_Liu_2017,Peng_Luo_Li_Liu_Zheng_2018} proposed techniques for detecting anomalous nodes in graphs based on matrix decomposition using homophily assumption, which states that connected nodes have similar features. Moreover, the exploration of deep learning \cite{Peng_Luo_Li_Liu_Zheng_2018,Li_Huang_Li_Du_Zou_2019,Bandyopadhyay_N_Vivek_Murty_2020} for graph anomaly detection is steadily increasing. DOMINANT \cite{Ding_Li_Bhanushali_Liu_2019} introduces GCN as a graph autoencoder to process both network structure and node attribute information. AnomalyDAE \cite{Fan_Zhang_Li_2020} uses GAT to encode network structure information. AEGIS \cite{ding2021inductive} introduces an unsupervised inductive anomaly detection method that can be applied to new nodes. 
 \cite{Chen_Liu_Wang_Dai_Lv_Bo_2020} proposed to use generative adversarial networks (GANs) \cite{Goodfellow_Pouget-Abadie_Mirza_Xu_Warde-Farley_Ozair_Courville_Bengio_2019} to generate anomalous nodes to support anomaly detection, while \cite{liu2021anomaly,xu2022contrastive,huang2023unsupervised} presented contrastive learning techniques for graph anomaly detection. 
\paragraph{\textbf{Graph Self-Supervised Learning}} Graph self-supervised learning (GSSL) \cite{lee2022augmentation} is an unsupervised approach that learns meaningful representations from graph data by constructing pretext tasks \cite{liu2022graph}. Three types of GSSL methods can be distinguished based on the different pretext tasks: contrastive, generative, and predictive. Contrastive methods generate multiple views for each graph instance and learn graph representations by contrasting the similarity and difference between different views \cite{You2020GraphCL,sun2019infograph,zhu2021graph,zeng2021contrastive}. Generative methods employ autoencoders to reconstruct parts of the input graph \cite{zhu2020self, manessi2021graph, he2022masked, hou2022graphmae}. Predictive methods \cite{wu2021self, jin2020self, peng2020self} use statistical analysis or expert knowledge to generate pseudo-labels for graph data and then design some prediction-based proxy tasks based on these pseudo-labels to learn graph representation.

\section{Problem Definition}
The primary focus of this work is to address the task of GAD in attributed networks. Following previous studies \cite{Ding_Li_Bhanushali_Liu_2019,liu2022bond}, we consider the unsupervised setting in this paper, \textit{i.e.}, learning without both node category labels and anomaly labels. An attributed network can be represented as $\mathcal{G}=(\mathcal{V}, \mathcal{E}, \bm{X})$, where $\mathcal{V}=\{v_1,\dots,v_n\}$ is the set of $n$ nodes, $\mathcal{E}$ is the set of $m$ edges, and $\bm{X} \in \mathbb{R}^{n \times d}$ is the attribute matrix. The structural information could also be represented by a binary adjacency matrix $\bm{A} \in \mathbb{R}^{n \times n}$. Specifically, $\bm{A}_{ij} = 1$ if there exists a connection between nodes $v_i$ and $v_j$, and $\bm{A}_{ij} = 0$ if not. The graph Laplacian matrix $\bm L$ is defined as $\bm D- \bm A$, where $\bm D$ is the degree matrix.

Given this attributed network, the aim of GAD is to identify nodes that deviate significantly from the majority in terms of both structural and attribute features. We attempt to formulate an anomaly function \cite{liu2022bond} that assigns an anomaly score to each node $v_{i}$. Nodes that exceed the predefined anomaly threshold $\lambda$ are classified as anomalous, while others are considered normal.

The anomalous nodes in the attributed graph can be categorized into two types \cite{ma2021comprehensive}:
\begin{itemize}
    \item \textbf{Structural anomalies} refer to densely connected nodes or other connection patterns that deviate from the sparsely connected regular nodes.
    \item \textbf{Contextual anomalies} are nodes whose attributes exhibit significant differences compared to their neighboring nodes.
\end{itemize}

\begin{figure*}[!t]
  \centering
  \includegraphics[width=0.92\textwidth]{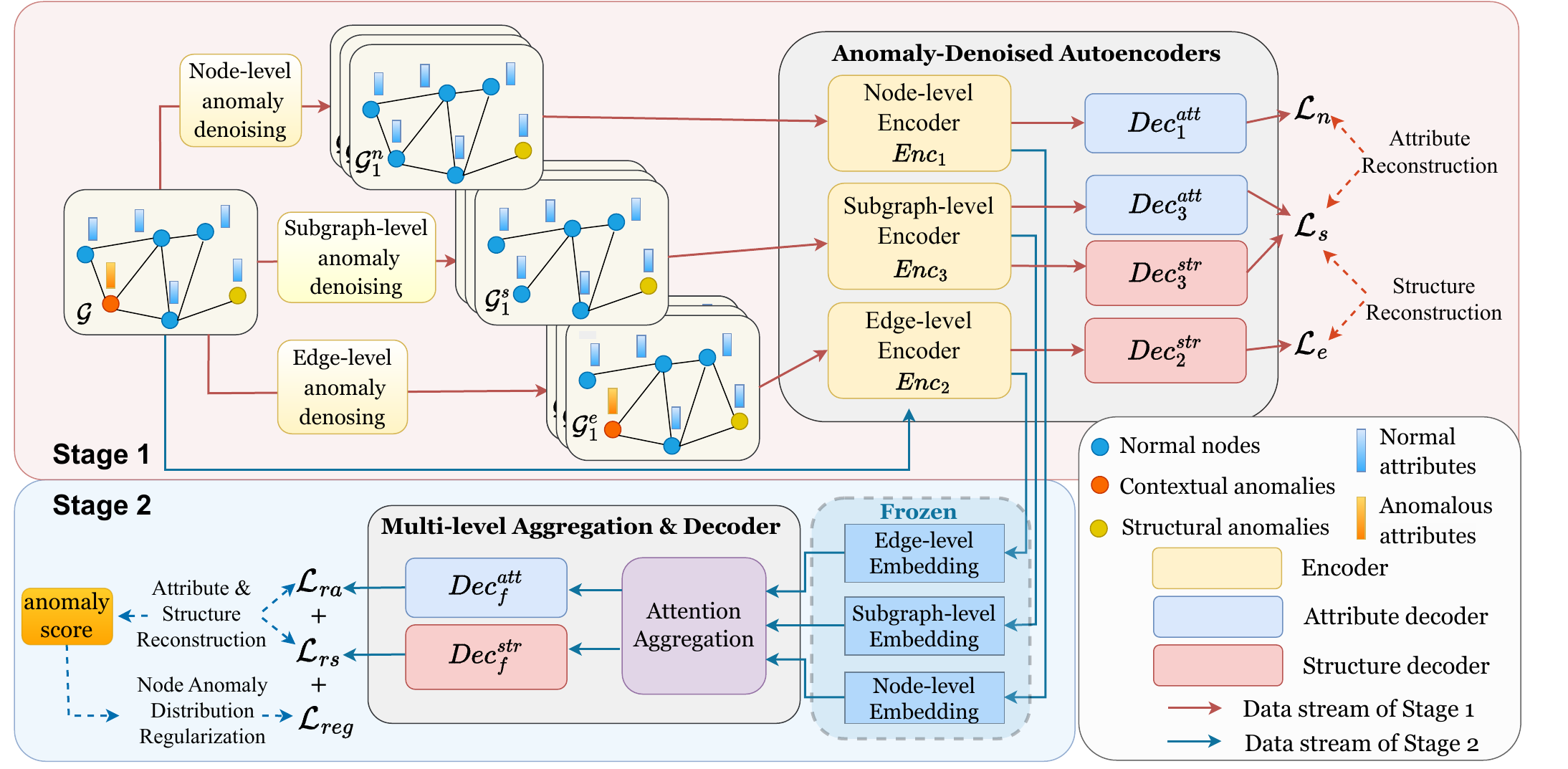}
 \caption{Overall framework of ADA-GAD. In Stage 1, we pretrain the graph autoencoders at three levels using the anomaly-denoised augmentation to mitigate the negative impact of anomalous patterns in the graph. In Stage 2, we retrain decoders based on multi-level embeddings obtained from fixed encoders, together with a regularization to sharpen the anomaly score's distribution.}
\label{fig:framework1}
\end{figure*}

\section{Methodology}
Previous reconstruction-based GAD models usually consists of a graph encoder and two graph decoders. Specifically, the attribute decoder reconstructs the node attributes, and the structural decoder reconstructs the adjacency matrix. The resulting reconstruction errors from both decoders are combined to calculate anomaly scores for the nodes. These anomaly scores are then ranked, and nodes with higher scores are identified as anomalies.

As discussed in the Introduction, directly reconstructing the original graph containing mixed anomalies will suffer from \textit{Anomaly Overfitting} and \textit{Homophily Trap}, degenerating the detection performance. Ideally, training the graph autoencoders on the graph with fewer anomalies and utilizing it for detection is the best way to address this issue. However, this is infeasible in the unsupervised detection setting due to the absence of ground-truth anomaly information. Instead, we resort to an anomaly-denoised pretraining process which reduces the anomaly rate of the graph by augmentation and pretrains via the reconstruction on the anomaly-denoised graph. After mitigating the negative impact of anomalies on the encoder, we freeze it and only retrain the decoder on the original graph before proceeding with the subsequent detection. This forms a two-stage framework in Figure \ref{fig:framework1}.

\subsection{Stage 1: Anomaly-Denoised Pretraining}

This stage generates the anomaly-denoised graphs and pretrains the graph autoencoders on them so that the auto-encoders can focus on the normal patterns. It paves the way for the subsequent anomaly detection stage by increasing the reconstruction error of anomalous nodes.

For the anomaly-denoised augmentation, we need to ensure that the anomaly level of the augmented graph is lower than that of the original graph. Then a key question arises: how to quantify the anomaly level of a graph? Some prior researches have shown that the level of anomaly in a signal $y$ on the graph $\mathcal{G}$ relates to its spectral statistics such as High-frequency Area Energy $E_{\text{high}}$ \cite{tang2022rethinking,gao2023addressing}
\begin{equation}
\label{eq:high_freq}
E_{\text{high}}(\mathcal{G},\bm{y})=\frac{\bm y^T\bm L \bm y}{\bm y^T\bm y}.
\end{equation}
As the anomaly ratio of signal $y$ in the graph $\mathcal{G}$ increases, $E_{\text{high}}(\mathcal{G},\bm{y})$ also increases, known as `right-shift' \cite{tang2022rethinking}.
Inspired by this, we define the anomaly degree at the attribute and structure level based on the corresponding characteristics as follows.

\begin{definition}[Attribute Anomaly Magnitude]
The attribu\linebreak -te anomaly magnitude on the graph $\mathcal{G}$ is defined as:
\begin{equation}
\label{eq:ano}
A_{\text{ano}}(\mathcal{G}) = E_{\text{high}}(\mathcal{G},\bm{X})=\frac{\bm{X}^T \bm{L} \bm{X}}{\bm{X}^T \bm{X}}.
\end{equation}
\end{definition}

\begin{definition}[Structural Anomaly Magnitude]
The structural anomaly magnitude on the graph $\mathcal{G}$ is defined as:
\begin{equation}
\label{eq:struct_freq}
S_{\text{ano}}(\mathcal{G})= E_{\text{high}}(\mathcal{G},\bm{D})= \frac{\bm{D}^T \bm{L} \bm{D}}{\bm{D}^T \bm{D}}.
\end{equation}
\end{definition}

A larger $A_{\text{ano}}$ for $\mathcal{G}$ indicates more pronounced variations in its attributes, leading to a higher ratio of contextual anomalies. And $S_{\text{ano}}$ characterizes the ratio of structural anomalies similarly.
\begin{definition}[Graph Anomaly Magnitude]
The anomaly magnitude on a graph $\mathcal{G}$ is defined as the sum of the attribute anomaly magnitude $A_{\text{ano}}$ and the structural anomaly magnitude $S_{\text{ano}}$:
\begin{equation}
\label{eq:graph_anomaly}
G_{\text{ano}}(\mathcal{G}) = A_{\text{ano}}(\mathcal{G}) + S_{\text{ano}}(\mathcal{G}) \nonumber.
\end{equation}
\end{definition}

Apparently, as the proportion of anomalous nodes within the graph increases, the graph anomaly magnitude also tends to rise. This provides a good measure for the anomaly-denoised augmentation process.
The objective of the augmentation is to minimize the anomaly level of the augmented graph $\mathcal{G}' = (\bm{X}', \bm{A}')$ within predefined augmentation budgets:
\begin{equation}
\label{eq:anomaly_denoised_augmentation_revised}
\begin{aligned}
& \underset{\substack{\bm{A}', \bm{X}'}}{\text{minimize}} \quad G_{\text{ano}}(\mathcal{G}'), \\
\text{subject to} \quad & \alpha \leq G_{\text{ano}}(\mathcal{G}) - G_{\text{ano}}(\mathcal{G}') \leq \beta, \\
& \begin{aligned}
\|\bm{X} - \bm{X}'\|^2_F \leq \epsilon_1, 
\|\bm{A} - \bm{A}'\|^2_F \leq \epsilon_2,
\end{aligned}
\end{aligned}
\end{equation}
where $\alpha>0$ and $\beta>0$ set the acceptable bounds for anomaly rate reduction, $\|\cdot\|_F$ denotes the Frobenius norm. Additionally, $\epsilon_1$ and $\epsilon_2$ are both small and denote the augmentation budget for the attribute and structure, respectively.

However, because the graph's adjacency matrix is discrete, directly solving this optimization problem is very challenging \cite{zhu2021survey}. Thus, we turn to design a learning-free augmentation approach to achieve our goal. We introduce a simplified graph masking strategy to generate the augmented graph and conduct denoising pretraining at three levels: node-level, edge-level, and subgraph-level.

\subsubsection{Node-level denoising pretraining}
For node-level anomaly denoising, we randomly select a subset of nodes $\mathcal{V}^n\subset\mathcal{V}$ for replacement-based masking using a probability of $p_r$. The masked node features are adjusted as follows:
\begin{equation}
\widetilde{\bm{x}}_{i}=
\begin{cases}
\bm{x}_j & v_i\in\mathcal{V}^n \\
\bm{x}_i & v_i\notin \mathcal{V}^n,
\end{cases}
\end{equation}
where we randomly choose another node (denoted as $j$) and replace the original feature $\bm{x}_i$ with $\bm{x}_j$ if $v_i\in\mathcal{V}^n$. Additionally, we also introduce a probabilistic mechanism where the feature of each node $v_i\in\mathcal{V}^n$ is randomly transited to zero with a probability of $p_z$. After the augmentation, we could calculate the corresponding graph anomaly magnitude and check if it satisfies the constraint in Problem \eqref{eq:anomaly_denoised_augmentation_revised}. The augmented graph is valid if the condition holds.

We repeat the above augmentation steps multiple times, generating a collection of valid augmented graphs of length $l_n$, denoted as $\mathcal{C}_n = \{\mathcal{G}_1^n, \mathcal{G}_2^n, ..., \mathcal{G}_{l_n}^n\}$, where each $\mathcal{G}_k^n=(\mathcal{V}, \mathcal{E}, \bm{X}_k^n)$ satisfies $G_{\text{ano}}(\mathcal{G}_k^n) \leq \theta$, $\bm{X}_k^n$ is the masked attribute matrix generated each time, $\theta = G_{\text{ano}}(\mathcal{G}) -\alpha$ is the anomaly degree threshold.

For each $\mathcal{G}_k^n$, we feed it into the node-level graph autoencoders consisting of the GNN encoder $Enc_1$ and the attribute decoder $Dec_1^{att}$ \cite{Ding_Li_Bhanushali_Liu_2019,Fan_Zhang_Li_2020}, and then obtain the reconstructed features:
\begin{equation}
\bm{\hat{\bm{X}}}_k^n=Dec_1^{att}(Enc_1(\mathcal{G}^n_k)).
\end{equation}
In the pretraining, the autoencoder should especially learn to reconstruct the feature $\bm{x}_i$ of the masked node $v_i \in \mathcal{V}$. Then the node-level reconstruction loss $\mathcal{L}_n$ is the sum of reconstruction losses over $\mathcal{C}_n$:
\begin{equation}\label{eq:node-level}
    \mathcal{L}_{n} = \sum^{l_n}_{k=1} \| \bm{X}_k^n- \hat{\bm{X}}_k^n \|_{F}^2.
\end{equation}

\subsubsection{Edge-level denoising pretraining}

Similar to the node-level pretraining, we randomly select a subset of edges $\mathcal{E}^e$ from $\mathcal{E}$ and apply masking with a probability of $q$, resulting in the corresponding entries in the adjacency matrix being set to zero. After multiple times of augmentation, we obtain a collection of $l_e$ edge masking graphs, denoted as $\mathcal{C}_e = \{\mathcal{G}_1^e, \mathcal{G}_2^e, ..., \mathcal{G}_{l_e}^e\}$, where each $\mathcal{G}_k^e =(\mathcal{V}, \mathcal{E} \setminus \mathcal{E}^e_k, \bm{X})$ also fulfills the condition $G_{\text{ano}}(\mathcal{G}_k^e) \leq \theta$, $\mathcal{E}^e_k$ is edge subset generated each time.

The edge-level autoencoders take the edge-level masked graph $\mathcal{G}_k^e$ as input and aim to reconstruct the denoised graph structure $\mathcal{E} \setminus \mathcal{E}^e_k$. The reconstructed adjacency matrix for each augmented $\bm{A}_k^e$ can be expressed as:
\begin{equation}
    \bm{\hat{\bm{A}}}_k^e=Dec_2^{str}(Enc_2(\mathcal{G}^e_k)),
\end{equation}
where $Enc_2$ and $Dec_2^{str}$ denote the GNN encoder and the structural decoder, respectively.

The loss function of the edge-level autoencoders $\mathcal{L}_{\text{e}} $ can be defined as:
\begin{equation}\label{eq:edge-level}
    \mathcal{L}_{e} =\sum^{l_e}_{k=1} \| \bm{A}_k^e- \hat{\bm{A}}_k^e \|_{F}^2.
\end{equation} 

\subsubsection{Subgraph-level denosing pretraining}
In addition, we propose a novel pretext task of subgraph masking pretraining. We employ random walk-based subgraph sampling for masking, adopting similar node and edge masking strategies as the above masking processes. In the resulted augmented graph sets $\mathcal{C}_s = \{\mathcal{G}_1^s, \mathcal{G}_2^s, ..., \mathcal{G}_{l_s}^s\}$, each $\mathcal{G}_k^s =(\mathcal{V}, \mathcal{E} \setminus \mathcal{E}^s_k, \bm{X}_k^s)$ satisfies $G_{\text{ano}}(\mathcal{G}_k^s) \leq \theta$, where $\mathcal{E}^s_k$ is edge subset generated each time, $\bm{X}_k^s$ is attribute matrix generated each time.

Subgraph-level masking can be viewed as a specific combination of node- and edge-level masking. The reconstructed feature and structure are:
\begin{equation}
    \begin{aligned}
        \hat{\bm{A}}^s_k&=Dec_3^{att}(Enc_3(\mathcal{G}_k^s)), \\
        \bm{\hat{X}}^s_k&=Dec_3^{str}(Enc_3(\mathcal{G}_k^s)),
    \end{aligned}
\end{equation}
where $Enc_3$ is a GNN encoder, $Dec_3^{att}$ and $Dec_3^{str}$ denote the attribute and structure decoder, respectively. The corresponding reconstruction loss function can be formulated as follows:
\begin{equation}
\mathcal{L}_{s} = \mathcal{L}_{sn}+\mathcal{L}_{se},
\end{equation} 
where $\mathcal{L}_{sn}$ and $\mathcal{L}_{se}$ are computed by substituting $\bm{\hat{X}}^s_k$ and $\hat{\bm{A}}^s_k$ into Eq.~\eqref{eq:node-level} and \eqref{eq:edge-level}, respectively.

In Stage 1, all the above three procedures are simultaneously adopted to pretrain the corresponding autoencoders. The various levels of denoising pretraining help the model discover the underlying normal node patterns.

\subsection{Stage 2: Retraining for Detection}
In this stage, the graph is no longer masked for learning. We fix the pretrained graph encoders, discard the pretrained decoders and retrain two unified decoders (one for attribute and another for structure) from scratch to detect anomalous information in the graph.

\subsubsection{Multi-level embedding aggregation}
The pretrained encoders produce embeddings at three levels. They are firstly passed through a fully connected layer, and then aggregated using an attention mechanism. This yields an aggregated multi-level embedding denoted as $\bm{h}$ for each node. 

\subsubsection{Graph reconstruction for anomaly detection}\label{sec:graph_rec}
The aggregated multi-level embeddings $\bm{h}$ are fed into an attribute decoder $Dec^{att}_f$ and a structure decoder $Dec_{f}^{str}$ for reconstruction. Specifically, we reconstruct the adjacency matrix $\bm{A}$ and attribute matrix $\bm{X}$ of the original graph as $\hat{\bm{A}}$ and $\hat{\bm{X}}$, respectively. The corresponding graph reconstruction loss $\mathcal{L}_{rec}$ is:
\begin{equation}
    \begin{aligned} \mathcal{L}_{rec} & =(1-\gamma) \mathcal{L}_{rs}+\gamma \mathcal{L}_{ra} \\ & =(1-\gamma)\|\bm{A}-\hat{\bm{A}}\|_F^2+\gamma\|\bm{X}-\hat{\bm{X}}\|_F^2,
    \end{aligned}
\end{equation}
where $\gamma\in[0,1]$ is a balance hyperparameter. And the anomaly score $s_i$ for the $i$-th node is defined as:
\begin{equation}
           s_i=(1-\gamma)\left\|\bm{a}_i-\hat{\bm{a}}_i\right\|_2+\gamma\left\|\bm{x}_i-\hat{\bm{x}}_i\right\|_2.
\end{equation}
where $\hat{\bm{a}}_i$ and $\bm{a}_i$ represent the reconstructed and original structure vector of node $v_i$, respectively. Similarly, $\hat{\bm{x}}_i$ and $\bm{x}_i$ are the $i$-th reconstructed and original attribute vector, respectively.

\subsubsection{Node Anomaly Distribution Regularization}
We propose a novel approach to regularize the model with a node anomaly distribution loss $\mathcal{L}_s$ that enforces sparsity on the anomaly distribution, further mitigating the \textit{Anomaly Overfitting}. Considering that overfitting occurs when all nodes are reconstructed very well, we intentionally introduce some non-uniformity in the anomaly distribution around nodes to enhance the difficulty of reconstruction. Therefore, we require the following anomaly distribution $\mathcal{A}_{i}$ of node $v_i$, \textit{e.g.}, the anomaly scores of a node and its neighbors, to be sharper:
\begin{equation}
    \mathcal{A}_{i}=\frac{s_i^{- \tau}}{\sum_{j \in \mathcal{N}_i} s_j^{-\tau}},
\end{equation}
where $\mathcal{N}_i$ represents the neighborhood of node $v_i$, and $\tau\in(0,1)$ is a temperature coefficient. Then the corresponding entropy $\mathcal{S}_i$ is:
\begin{equation}
    \begin{aligned}\mathcal{S}_{i} & =- \mathcal{A}_{i}\log \mathcal{A}_{i} \\
    &=\frac{s_i^{- \tau}}{\sum_{j \in \mathcal{N}_i} s_j^{-\tau}} (\log \sum_{j \in \mathcal{N}_i} s_j^{-\tau} -\tau \log s_{i}).
    \end{aligned}
\end{equation}  
In fact, $\mathcal{S}_i$ represents the smoothness level of the anomaly distribution around node $i$. A higher value of $\mathcal{S}_i$ indicates a sharper anomaly distribution. Accordingly, the node anomaly distribution regularization term $\mathcal{L}_{reg}$ is defined as: 
\begin{equation}
    \mathcal{L}_{reg}=- \sum_{v_i \in \mathcal{V}} \mathcal{S}_{i}.
\end{equation}

\subsubsection{Optimization Objective}

Putting all together, we have the overall loss function in this stage:
\begin{equation}
    \begin{aligned}
    \mathcal{L} &=\mathcal{L}_{rec}+\mathcal{L}_{reg} \\
    &=(1-\gamma) \mathcal{L}_{ra}+\gamma \mathcal{L}_{rs}+\gamma_{reg}\mathcal{L}_{reg},
    \end{aligned}
\end{equation}
where $\gamma_{reg}$ is a weight hyperparameter that should be small to avoid influencing optimization of $\mathcal{L}_{rec}$. Based on this loss, we retrain the aggregation and the decoder modules. After retraining, we sort the nodes based on their anomaly scores $s_i$, and take the nodes with higher $s_i$ as anomalous nodes according to the given anomaly rate.

\section{Experiments} \label{sec:eval}

\subsection{Experimental Setup} \label{ssec:exp-set}

\subsubsection{Datasets}

\begin{table}[!t]\small
    \centering  
    \caption{Statistics of dataset \textmd{($*$ indicates the dataset with injected anomalies)}.}\label{table:dataset}
    \setlength{\tabcolsep}{5pt}
\resizebox{0.41\textwidth}{!}
{\begin{tabular}{lccccc}
\midrule
\textbf{Dataset} & \textbf{Nodes} & \textbf{Edges} & \textbf{Feat} & \textbf{Anomalies} & \textbf{Ratio} \\ \midrule
Cora$^{*}$            & 2,708          & 11,060         & 1,433         & 138                & 5.1\%          \\
Amazon$^{*}$           & 13,752         & 515,042        & 767           & 694                & 5.0\%          \\
Weibo            & 8,405          & 407,963        & 400           & 868                & 10.3\%         \\
Reddit           & 10,984         & 168,016        & 64            & 366                & 3.3\%          \\
Disney           & 124            & 335            & 28            & 6                  & 4.8\%          \\
Books            & 1,418          & 3,695          & 21            & 28                 & 2.0\%          \\
Enron            & 13,533         & 176,987        & 18            & 5                  & 0.4\%          \\ \midrule
\end{tabular}}
\end{table}

We conducted experiments on two datasets injected with synthetic anomalies: Cora \cite{sen2008collective}, Amazon \cite{Shchur_Mumme_Bojchevski_Günnemann_2018}, and five manually labeled datasets with anomalies: Weibo \cite{Zhao_Deng_Yu_Jiang_Wang_Jiang_2020}, Reddit \cite{Kumar_Zhang_Leskovec_2019}, Disney \cite{Müller_Sánchez_Mulle_Böhm_2013}, Books \cite{Sánchez_Müller_Laforet_Keller_Böhm_2013}, and Enron \cite{Sánchez_Müller_Laforet_Keller_Böhm_2013}. We injected contextual anomalies into datasets with no labeled anomalies by swapping node attributes, and structural anomalies by altering node connections within the graph, maintaining an equal number for each type in alignment with prior research \cite{Ding_Li_Bhanushali_Liu_2019, liu2022pygod}. The statistics of all the datasets are recorded in Table \ref{table:dataset}.

\begin{table*}[!t]\small
    \centering
    \renewcommand\arraystretch{1.1}
    \setlength{\tabcolsep}{5pt}
    \caption{AUC (\%) results (mean $\pm$ std). The best result is shown in \textbf{bold}, while the second best is marked with \underline{underline}. }\label{table:main_result}
\resizebox{0.9\textwidth}{!}{
\begin{tabular}{cllc|ccccccc}
\toprule[1pt]
\midrule
\multicolumn{4}{c|}{\multirow{1}{*}{Algorithm}} & Cora & Amazon & Weibo & Reddit & Disney & Books & Enron \\
\midrule
\multicolumn{3}{c|}{}                           & SCAN                                             & 64.95±0.00                     & 65.85±0.00                     & {\color[HTML]{333333} 70.63±0.00} & 49.67±0.00                     & 50.85±0.00                     & 52.42±0.00                     & 53.70±0.00                     \\
\multicolumn{3}{c|}{}                           & Radar                                            & 53.28±0.00                     & 58.93±0.00                     & \underline{98.27±0.00}              & \underline{56.64±0.00}                     & 50.14±0.00                     & 56.21±0.00                     & 64.10±0.00                     \\
\multicolumn{3}{c|}{\multirow{-3}{*}{Non-Deep}} & ANOMALOUS                                        & 35.13±1.07                     & 71.49±1.71                     & \underline{98.27±0.00}              & 51.58±8.66                     & 50.14±0.00                     & 52.51±0.00                     & 63.65±0.36                     \\ \midrule
\multicolumn{3}{c|}{}                           & MLPAE                                            & 70.91±0.07                     & 74.20±0.00                     & 90.01±0.42                        & 49.74±1.70                     & 48.02±0.00 & 51.28±5.36 &41.55±2.59 \\
\multicolumn{3}{c|}{}                           & GCNAE                                            & 70.90±0.00                     & 74.20±0.00                     & 88.98±0.31                        & 50.70±0.46                     & 47.34±1.31                     & 54.81±1.59                     & 66.86±0.54                     \\
\multicolumn{3}{c|}{}                           & DOMINANT                                         & 76.71±0.07                     & 74.20±0.00                     & 92.17±0.41                       & 56.20±0.06                     & 52.91±3.04                     & 40.14±2.66                     & 54.93±0.66                     \\
\multicolumn{3}{c|}{}                           & DONE                                             & 83.60±1.45                     & 73.38±4.37                     & 86.86±0.38                        & 51.40±2.26                     & 48.42±4.23                     & 54.05±1.64                     & 61.07±3.15                     \\
\multicolumn{3}{c|}{}                           & AdONE                                            & 82.12±0.71                     & 79.31±2.60                     & 82.98±0.63                        & 51.53±1.38                     & 50.93±2.34                     & 54.13±1.60                     & 58.36±7.34                     \\
\multicolumn{3}{c|}{}                           & AnomalyDAE                                       & 80.99±0.07                     & 77.39±0.01                     & 92.99±0.44                        & 52.21±2.03                     & 48.29±4.17                     & 59.86±4.82                     & 45.85±13.16                     \\
\multicolumn{3}{c|}{}                           & GAAN                                             & 68.32±1.38                     & 77.70±0.34                     & 92.53±0.01                        & 51.23±1.19                     & 48.02±0.00                     & 53.38±2.13                     & 56.55±11.69                    \\
\multicolumn{3}{c|}{}                           & CoLA                                             & 56.88±1.63                    & 61.00±1.09                     & 22.18±3.36                        & 53.21±1.10                     & 54.46±7.67                     & 49.69±4.20                     & 58.53±9.38                     \\
\multicolumn{3}{c|}{}                           & OCGNN                                            & 50.02±0.14                     & 49.99±0.04                     & 79.68±5.76                        & 48.76±3.57                     & 68.19±1.45                     & 57.33±4.14                     & 54.39±6.11                     \\
\multicolumn{3}{c|}{}                           & CONAD                                            & 84.34±0.03                     & \underline{82.62±0.26}                     & 90.87±0.59                        & 56.02±0.01                     & 45.38±4.64                     & 40.82±1.18                     & 54.67±0.48                     \\ \cmidrule{4-11} 
\multicolumn{3}{c|}{}                           & ADA-GAD$_{rand}$ & 81.61±0.01 & 76.36±0.12 &90.74±0.65    & 56.03±0.38 & 68.56±2.94 & 61.75±2.20 & 66.12±4.87 \\
\multicolumn{3}{c|}{}                           & ADA-GAD$_{node}$                      & 84.13±0.02                     & 77.38±0.02                     & 96.39±0.74                        & 56.33±0.16                     & 68.05±3.70                     & \underline{62.77±2.31}                     & 71.55±2.27                     \\
\multicolumn{3}{c|}{}                           & ADA-GAD$_{edge}$                      & 84.10±0.01                     & 81.85±0.03                     & 94.52±0.70                        & 56.37±0.10                     & 68.42±3.19                     & 62.71±2.17                     & 72.34±1.42                     \\
\multicolumn{3}{c|}{}                           & ADA-GAD$_{subgraph}$                      & \underline{84.42±0.01}                     & 81.79±0.04                     & 96.69±0.59                        & 55.58±0.36                     & \underline{68.59±2.62}                     & 62.70±2.13                     & \underline{72.86±0.88}                     \\ \cmidrule{4-11} 
\multicolumn{3}{c|}{\multirow{-15}{*}{Deep}}    & ADA-GAD                                         & \textbf{84.73±0.01}            & \textbf{83.25±0.03}            & \textbf{98.44±0.33}                       & \textbf{56.89±0.01}            & \textbf{70.04±3.08}            & \textbf{65.24±3.17}            & \textbf{72.89±0.86}            \\ \midrule
\bottomrule[1pt]
\end{tabular}}
\end{table*}

\subsubsection{Competitors}\label{sec:baseline} We adopt three non-deep learning methods for graph anomaly detection comparison: 
 \textbf{SCAN} \cite{Xu_Yuruk_Feng_Schweiger_2007},
  \textbf{Radar} \cite{Li_Dani_Hu_Liu_2017},
 \textbf{ANOMALOUS} \cite{Peng_Luo_Li_Liu_Zheng_2018}.
Additionally, we have also selected the following deep learning-based competitors: \textbf{GCNAE} \cite{kipf2016variational}, \textbf{DOMINANT} \cite{Ding_Li_Bhanushali_Liu_2019}, \textbf{DONE} \cite{Bandyopadhyay_N_Vivek_Murty_2020}, \textbf{AdONE} \cite{Bandyopadhyay_N_Vivek_Murty_2020}, \textbf{AnomalyDAE} \cite{Sakurada_Yairi_2015}, \textbf{GAAN} \cite{Chen_Liu_Wang_Dai_Lv_Bo_2020}, \textbf{CoLA} \cite{liu2021anomaly}, \textbf{OCGNN} \cite{wang2021one}, \textbf{CONAD} \cite{xu2022contrastive}.

Meanwhile, we also implement four variants of the proposed ADA-GAD method to verify the effectiveness of the anomaly-denoised augmentation: \textbf{ADA-GAD$_{rand}$} refers to using random augmentation; \textbf{ADA-GAD$_{node}$, ADA-GAD$_{edge}$}, and \textbf{ADA-GAD$_{subgraph}$} utilize a single level of anomaly-denoised augmentation, denoising pretrained at only the node, edge, and subgraph level, respectively.

\subsubsection{Implementation Details}
We implement all the competitors with the PyGOD toolbox \cite{liu2022pygod}. We set the number of epochs/dropout rate/weight decay to 100/0.1/0.01, respectively. The embedding dimension $d$ is set to 12 for the Disney, Books, and Enron datasets, and 64 for the others. 

Our ADA-GAD method utilizes GCN as the encoders and decoders, except for the Enron and Weibo datasets, where we adopt GAT as the encoders and GCN as decoders. For the real-world datasets Disney, Books, and Enron, the encoder depth is set to 2 and the decoder depth is 1. For the other datasets, encoder and decoder depths are set to 1. During augmentation, the number of masks for nodes and edges is set within the range of 1 to 20, respectively. The number of random walks and walk length for the subgraph mask are both set to 2. $l_n$, $l_e$, and $l_s$ are all set to 10, with $\theta$ is assigned to the smallest $G_{\text{ano}}$ among $N\_{aug}$ random augmentations. In the experiments, $N\_{aug}$ is set to 30. The pre-training epoch and the retain epoch are both set to 20. AUC (Area under the ROC Curve) \cite{bradley1997use} is used as the performance metric.
We repeat all experiments 10 times using 10 different seeds.

\subsection{Performance Comparison}  \label{ssec:exp-sota}

All the experimental results are reported in Table \ref{table:main_result} reports all the experimental results. From the results, we have the following observations: (1) ADA-GAD consistently exhibits better AUC performance than other competitors, which validates the effectiveness of the proposed method. (2) A single anomaly-denoised pretraining branch is a little inferior to the combination of three-level anomaly-denoised pretraining branches but outperforms the random one. This phenomenon indicates that our anomaly-denoised training strategy successfully utilizes the information at the node, edge, and subgraph levels for anomaly detection. (3) On four datasets with relatively small feature dimensions (\textit{i.e.}, Reddit, Disney, Books, and Enron), some competitors might achieve poor AUC performances, which is consistent with the empirical results in the benchmark \cite{liu2022bond}. In contrast, our ADA-GAD demonstrates consistent improvement over the competitors, which again validates our motivation. (4) The non-deep learning methods, Radar and ANOMALOUS, outperform the other deep learning competitors on the Weibo and Reddit datasets. This counter-intuitive result indicates that these deep learning methods might suffer from severe over-fitting. As a comparison, the proposed method utilizes the denoised augmentation of real-world datasets and takes anomaly-denoised graphs as inputs, which forces the model to pay more attention to normal patterns during pretraining and thus overcomes the over-fitting issue.

\begin{figure}[t]
    \centering
    \begin{minipage}[t]{0.2\textwidth}
        \includegraphics[width=\textwidth]{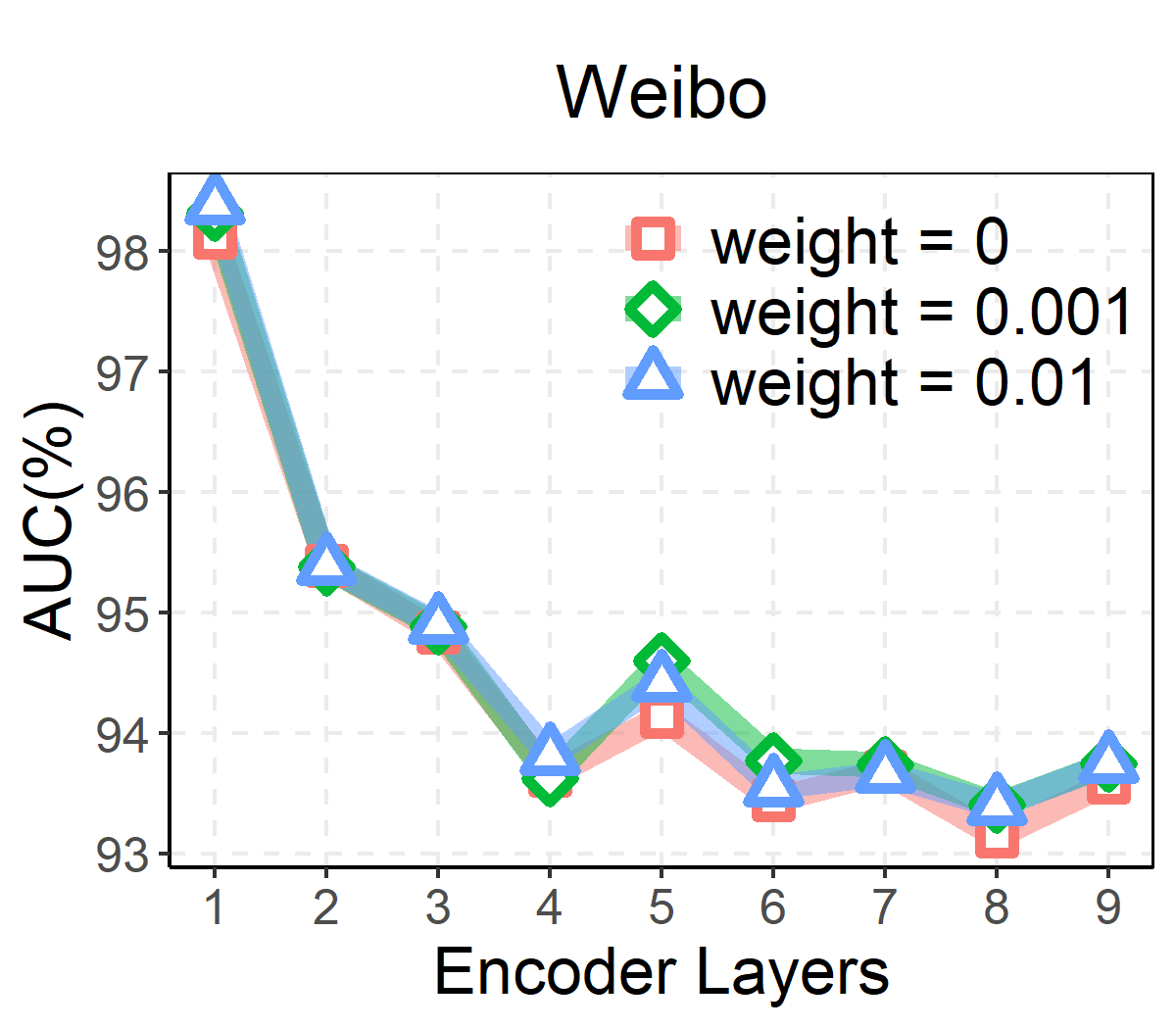}
    \end{minipage}
    \begin{minipage}[t]{0.2\textwidth}
    \includegraphics[width=\textwidth]{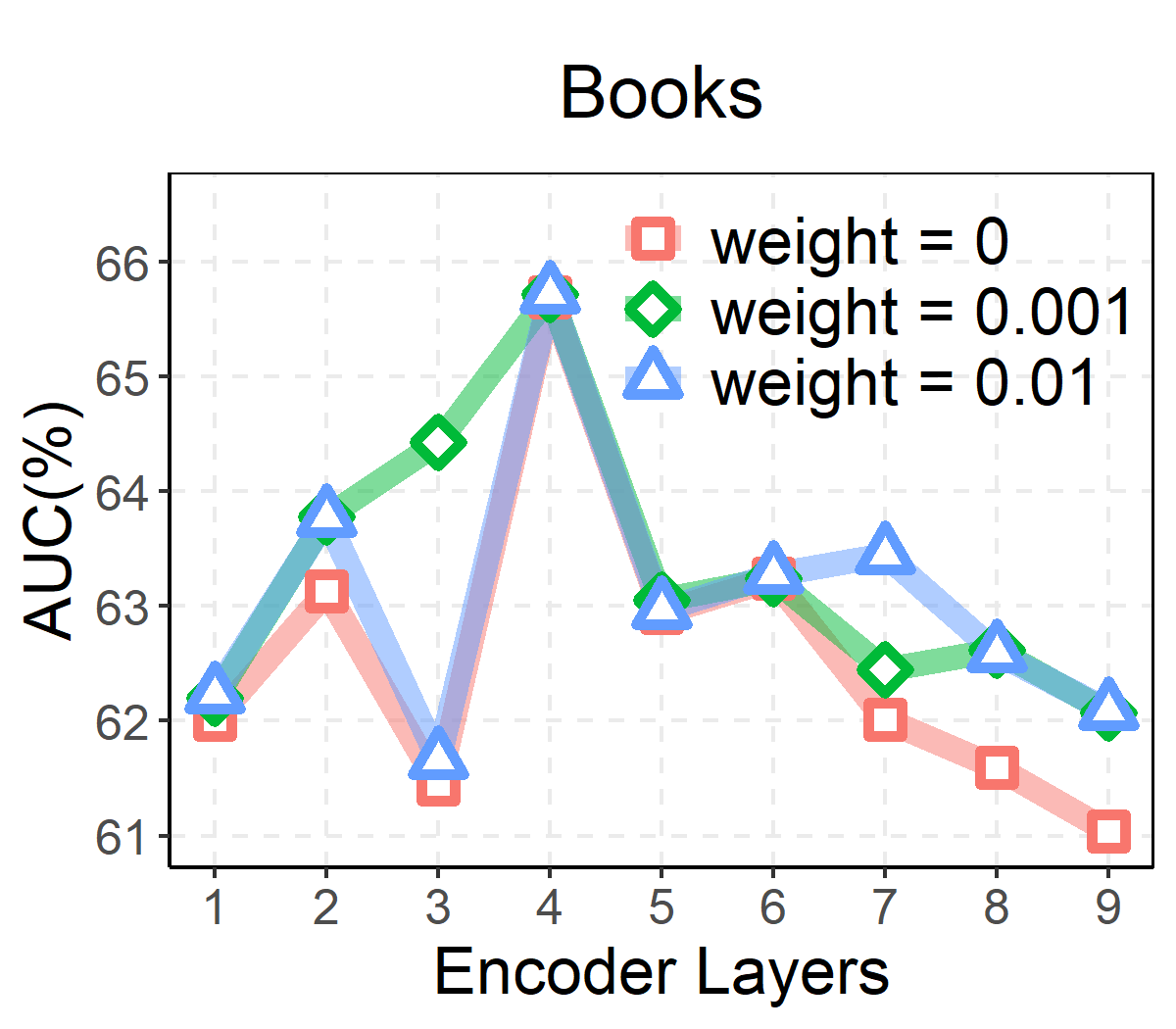}

    \end{minipage}
        \begin{minipage}[t]{0.2\textwidth}
        \includegraphics[width=\textwidth]{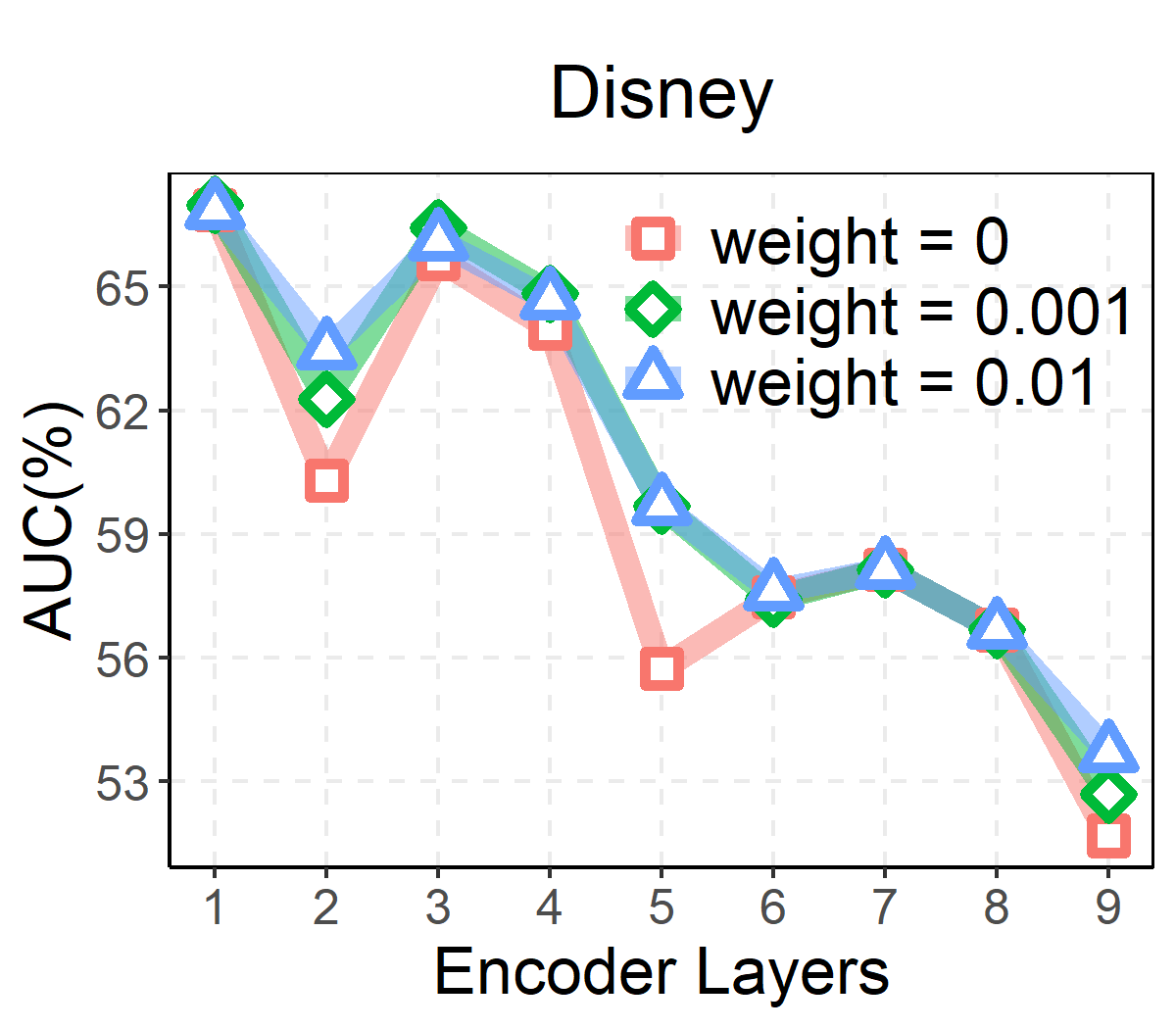}
    \end{minipage}
    \begin{minipage}[t]{0.2\textwidth}
    \includegraphics[width=\textwidth]{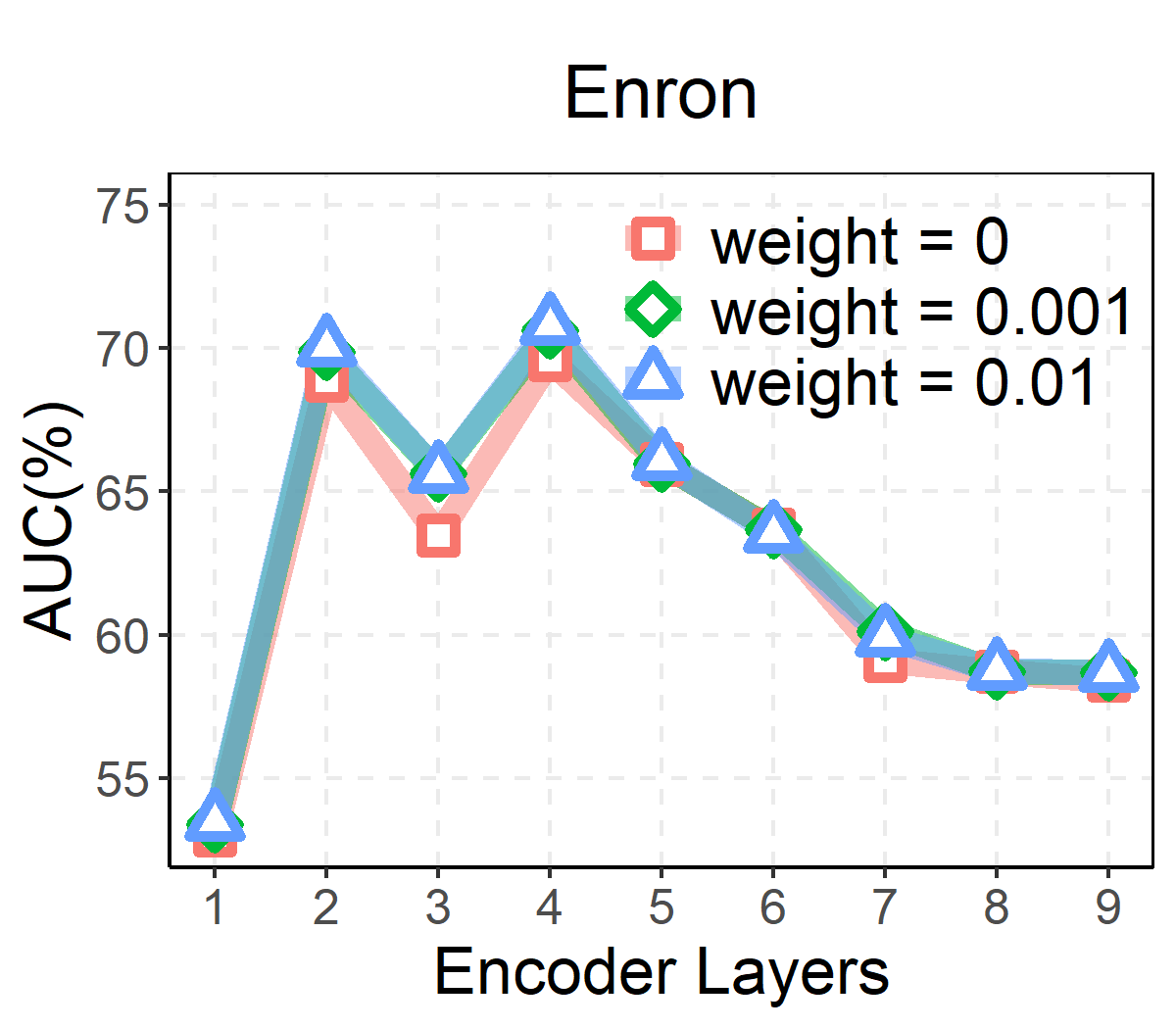}
    \end{minipage}
    \caption{Effect of the encoder depth and weight of node anomaly distribution regularization on four organic datasets.}
    \label{fig:ablation}
\end{figure}

\subsection{Ablation Study} 
\label{sec:abstudy}

\begin{figure}[t]
  \centering
  \includegraphics[width=0.34\textwidth]{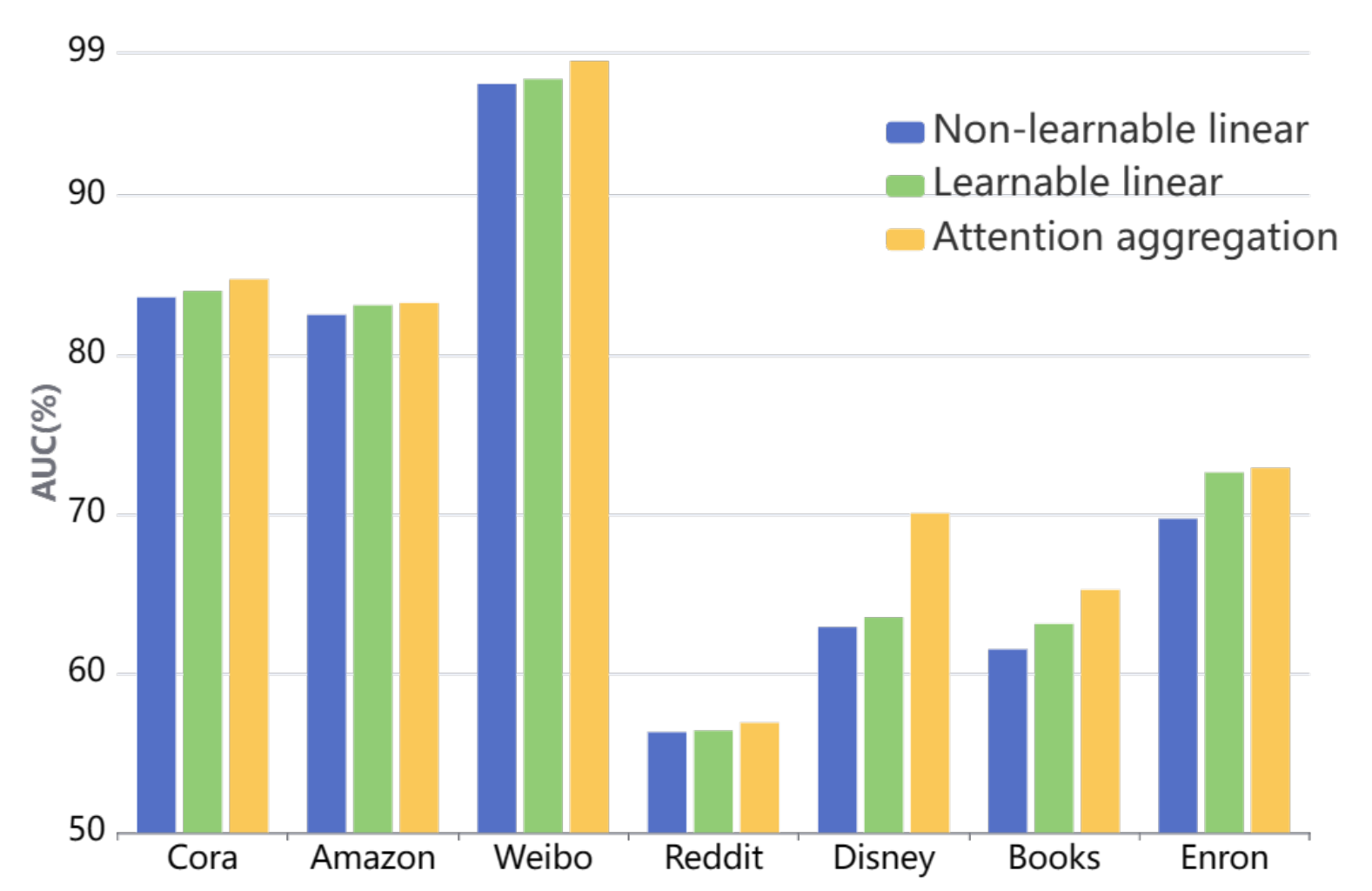}
  \caption{Performance comparison using different aggregation strategies.}
  \label{fig:aggr} 
\end{figure}

 \paragraph{Studies on Aggregation Strategies} We explored different aggregation strategies: non-learnable linear, learnable linear, and our attention aggregation. The non-learnable method uses fixed weights, whereas the learnable method optimizes weights through gradient descent. Figure \ref{fig:aggr} demonstrates the superior performance of our attention aggregation, highlighting its enhanced efficacy.

\paragraph{Studies on the Model Depth and Node Anomaly Distribution Regularization} To investigate the effectiveness of our model at different network depths, we evaluate the performance of the encoder with the number of layers ranging from 1 to 9 under different weights (0, 0.01, 0.001) of the node anomaly distribution regularization. As shown in Figure \ref{fig:ablation}, we can find that (1) the optimal number of encoder layers varies across datasets, with the Weibo dataset having an optimal number of 1 and the other three real-world datasets having an optimal number of 3 or 4. This suggests that the Weibo dataset is more prone to overfitting, consistent with our previous experimental findings. (2) After reaching the optimal number of layers, increasing the depth fails to improve performance. Fortunately, the node anomaly distribution regularization can alleviate this issue, as a larger weight within a small range can induce better performance. 

\section{Conclusion}
In this paper, we introduced ADA-GAD, a novel two-stage framework for graph anomaly detection. Through anomaly-denoised augmentation and a two-stage training framework, ADA-GAD effectively captures the normal patterns and enhances anomaly detection performance. Additionally, we introduce a node anomaly distribution regularization term to mitigate the model overfitting by constraining the anomaly distribution near nodes. Experimental results demonstrate that our proposed method achieves state-of-the-art performance on multiple benchmarks with injected or organic anomalous nodes. Future research can explore further refinements and extensions of ADA-GAD in different types of anomaly detection scenarios, such as supervised settings or graph-level anomaly detection.

\section{Acknowledgments}
This work was supported in part by the National Key R\&D Program of China under Grant 2018AAA0102000, in part by National Natural Science Foundation of China: 62236008, U21B2038, U23B2051, 61931008,  62122075 and 61976202, in part by the Fundamental Research Funds for the Central Universities, in part by Youth Innovation Promotion Association CAS, in part by the Strategic Priority Research Program of Chinese Academy of Sciences, Grant No. XDB28000000, in part by the Innovation Funding of ICT, CAS under Grant No.E000000, in part by the China National Postdoctoral Program under Grant GZB20230732, and in part by China Postdoctoral Science Foundation under Grant 2023M743441.

\bigskip

\bibliography{aaai24}

\section{Appendix} \label{appx}

\subsection{Signals in Graphs}
Given a graph $\mathcal{G}=(\mathcal{V}, \mathcal{E}, \bm{X})$, where $\mathcal{V}$ represents the set of vertices, $\mathcal{E}$ represents the set of edges, and $\bm{X}$ represents the node features, the normalized Laplacian matrix $\bm L$ is defined as $\bm D - \bm A$, where $\bm A$ is the adjacency matrix and $\bm D$ is the degree matrix with $\bm D_{ii}= \sum_{j} \bm A_{ij}$.

By performing the eigendecomposition on the Laplacian matrix $\bm L$, we obtain the eigenvalues arranged in ascending order as $0= \lambda_1 \leq \lambda_2 \leq \cdots \leq \lambda_n$ and their corresponding eigenvectors $\bm U=(\bm u_1, \bm u_2, \cdots, \bm u_n)$. This decomposition provides valuable insights into the graph's structural properties and spectral characteristics, forming the basis for various graph analysis and processing techniques.

We represent the signal on the graph $\mathcal{G}$ as $\bm y = (y_1, y_2,\cdots,y_n)^T \in \mathbb R^n$. Using the eigenvectors $\bm U$ of the Laplacian matrix as the basis functions for the graph Fourier transform, we obtain $\bm{\hat y}=\bm U^T \bm y=(\hat y_1, \hat y_2,\cdots,\hat y_N)^T$. Signals associated with high-frequency eigenvalues are referred to as high-frequency signals, whereas signals associated with low-frequency eigenvalues are considered as low-frequency signals on the graph.

\subsubsection{High-frequency Area Energy}

Based on previous studies \cite{tang2022rethinking,gao2023addressing}, the spectral energy $S_{ene}$ at $\lambda_k$ ($1 \leq k \leq N$) is given by:
\begin{equation}
S_{ene}(\mathcal{G},k)=\hat y_k^2/\sum_{i=1}^N \hat y_i^2.
\end{equation}

The high-frequency area energy $E_{\text{high}}$ concerning the signal $\bm{y}$ on graph $\mathcal{G}$ can be calculated as follows:

\begin{equation}
\label{eq:high_freq2}
E_{\text{high}}(\mathcal{G},\bm{y})=\frac{\sum_{k=1}^N\lambda_k \hat y^2_k}{\sum_{k=1}^N \hat y_k^2}=\frac{\bm y^T\bm L \bm y}{\bm y^T\bm y}.
\end{equation}

As the spectral energy of high-frequency signals $S_{ene}$ becomes concentrated in the larger eigenvalues, there is a corresponding increase within the high-frequency area energy $E_{\text{high}}$. Furthermore, due to the correlation between higher-frequency signals in the graph and fewer distinct patterns, an increase in the anomaly rate of signal $\bm y$ within the graph results in a corresponding energy increase in the high-frequency area $E_{\text{high}}(\mathcal{G},\bm{y})$. For detailed explanation and proof, please refer to \cite{tang2022rethinking}.

\subsection{Datasets Details}
\subsubsection{Datasets}

We conducted experiments on three datasets without manual labeling of anomalies:
\begin{itemize}
    \item \textbf{Cora} \cite{sen2008collective} is a collection of 2,708 computer science papers categorized into seven classes, with attributes such as titles, abstracts, and keywords used for node classification and graph neural network algorithm evaluation.
    \item \textbf{Amazon} \cite{Shchur_Mumme_Bojchevski_Günnemann_2018} is a product classification dataset extracted from the Amazon marketplace.
\end{itemize}
Following the approach of previous studies \cite{Ding_Li_Bhanushali_Liu_2019,liu2022pygod,Fan_Zhang_Li_2020}, we inject synthetic anomalous nodes into the Cora and Amazon datasets. Specifically, attribute anomalies are obtained by swapping some node attributes in the graph, while structural anomalies are obtained by changing some node connections.

Besides, we also conduct experiments on five manually labeled datasets (also known as organic datasets or real-world datasets \cite{liu2022bond}) with anomalies:
\begin{itemize}
    \item \textbf{Weibo} \cite{Zhao_Deng_Yu_Jiang_Wang_Jiang_2020} is a user-post-tag graph dataset from Tencent-Weibo, with suspicious users labeled based on temporal behavior.
    \item \textbf{Reddit} \cite{Kumar_Zhang_Leskovec_2019} is a user-subreddit graph dataset from the platform Reddit, with banned users labeled as anomalies.
    \item \textbf{Disney} \cite{Müller_Sánchez_Mulle_Böhm_2013} and \textbf{Books} \cite{Sánchez_Müller_Laforet_Keller_Böhm_2013} are co-purchasing networks of movies and books from Amazon, with anomalies labeled by high school students (Disney) or extracted from amazon information (Books).
    \item \textbf{Enron} \cite{Sánchez_Müller_Laforet_Keller_Böhm_2013} is an email network dataset with anomalies labeled based on identifying spam-sending email addresses.
\end{itemize}

\subsection{Competitors Details}
\paragraph{Non-deep learning methods} We adopt three for graph anomaly detection comparison:
\begin{itemize}
    \item \textbf{SCAN} \cite{Xu_Yuruk_Feng_Schweiger_2007}  detects clusters, hub nodes, and structural outliers
in a graph by analyzing its structure. 
    \item \textbf{Radar} \cite{Li_Dani_Hu_Liu_2017} detects outlier nodes by characterizing the residuals
of attribute information and its coherence with network information.
    \item \textbf{ANOMALOUS} \cite{Peng_Luo_Li_Liu_Zheng_2018}  performs joint anomaly detection and attribute selection to detect node anomalies based on the CUR decomposition and residual analysis, using both graph structure and node attribute information.
\end{itemize}

\paragraph{Deep learning methods} We also adopt the following deep learning-based anomaly detection methods as competitors:
\begin{itemize}
    \item \textbf{MLPAE} \cite{sakurada2014anomaly} employs a multiple layer perceptron (MLP) \cite{gardner1998artificial} as both an encoder and a decoder, utilizing a reconstruction-based approach for detecting and reconstructing node attributes.
    \item \textbf{GCNAE} \cite{kipf2016variational}  uses GCNs as both encoder and decoder to learn
node embeddings and reconstruct node attributes.
    \item \textbf{DOMINANT} \cite{Ding_Li_Bhanushali_Liu_2019} combines GCN and Autoencoders (AE) \cite{DBLP:journals/corr/abs-2003-05991} for outlier node
detection by using two-layer GCN as encoder, two-layer
GCN as decoder for attribute reconstruction, and one-layer
GCN and dot product as a structural decoder.
    \item \textbf{DONE} \cite{Bandyopadhyay_N_Vivek_Murty_2020}  leverages structural and attribute AE with MLP
for adjacency matrix and node attribute reconstruction.
    \item \textbf{AdONE} \cite{Bandyopadhyay_N_Vivek_Murty_2020} is a variant of DONE that uses an extra discriminator
for better alignment of embeddings.
    \item \textbf{AnomalyDAE} \cite{Fan_Zhang_Li_2020} utilizes structure AE and attribute AE
with node embeddings for outlier node detection.
    \item \textbf{GAAN} \cite{Chen_Liu_Wang_Dai_Lv_Bo_2020} is a GAN-based method that generates fake graphs
and identifies outlier nodes through node reconstruction
error and real-node identification confidence.
    \item \textbf{CoLA} \cite{liu2021anomaly}focuses on modeling the relations between nodes
and their neighboring subgraphs to detect community anomalies.
    \item  \textbf{OCGNN} \cite{wang2021one} combines the representation power of GNNs with the objective function of hypersphere learning to detect anomalous nodes. By learning the hyperspherical boundary, points outside the boundary are identified as outlier nodes.
    \item  \textbf{CONAD} \cite{xu2022contrastive} identifies anomalous nodes by leveraging prior human knowledge of different anomaly types. It does this by modeling the knowledge through data augmentation and integrating it into a Siamese graph neural network encoder using a contrastive loss. 
\end{itemize}

\subsection{Implementation Details}
\paragraph{Environment} The experiment utilized the following libraries and their respective versions: Python=3.7.0, CUDA\_version=11.7, torch=1.13, pytorch\_geometric=2.3.0, networkx=2.6.3, pygod=1.0.9.
\paragraph{Hardware configuration}The experiments were conducted on a Linux server with a single GeForce RTX 3090 GPU with 24GB memory.
\paragraph{Competitors}We implement all the competitors with the PyGOD toolbox \cite{liu2022pygod}. The embedding dimension is set to 12 for the Disney, Books, and Enron datasets, and 64 for the others. Other parameters of competitors are set to the default values provided by PyGOD.

\subsection{Details of Multi-level Embedding Aggregation}
Given the complete graph, we could obtain node-level, edge-level, and subgraph-level embeddings $\bm{e_1}$, $\bm{e_2}$, and $\bm{e_3}$ by the three encoders, respectively.

These embeddings focus on different levels of information on the graph. To aggregate this information, we design the following three aggregation mechanisms:
\begin{itemize}
    \item \textbf{Non-learnable linear aggregation}: We manually assign different scalar weights $\alpha_i$ to each embedding $\bm{e}_i (i=1,2,3)$ and sum them up to get the multi-level embedding $\bm{h}$:
    \begin{equation} \label{non-la}
        \bm{h}=\sum_{i=1}^3 {\alpha}_i \bm{e}_i,
    \end{equation}
    where the weight $\alpha_i$ subjects to $\sum_{i=1}^3\alpha_i=1,\ \alpha_i>0$. 
    
    \item \textbf{Learnable linear aggregation}: The scalar weight $\alpha_i$ is trainable in Equation (\ref{non-la}), which can be updated with gradient descent. 

    \item \textbf{Attention aggregation}: We pass the three-level embeddings through a fully connected network $fc$, followed by a softmax layer to obtain the weight vector $\bm{att}_{i}$ for each embedding $\bm{e}_i$:
    \begin{equation}
        \bm{att}_{i}=\frac{\exp( fc( \bm{e}_i ) )}{\sum_{j=1}^3 \exp(fc( \bm{e}_j ) )}.
    \end{equation}
    Then the weight $\bm{att}_i$ is adopted in Equation (\ref{non-la}) by replacing $\alpha_i$ with $\bm{att}_i$ and changing the weighted sum to the vector dot product.

\end{itemize}

    The results presented in the main text confirm the effectiveness of the Attention aggregation method, which outperforms others.


\end{document}